\pdfoutput=1

\documentclass[11pt]{article}

\usepackage[]{emnlp2021}

\usepackage{times}
\usepackage{latexsym}
\usepackage{flushend}
\usepackage[T1]{fontenc}

\usepackage[utf8]{inputenc}

\usepackage{microtype}

\usepackage{graphicx}

\usepackage{enumitem}

\usepackage{amsthm}

\title{Enhancing Legal Argument Mining with Domain Pre-training and Neural Networks}

\author{Gechuan Zhang$^1$, Paul Nulty$^2$, and David Lillis$^1$ \\
$^1$ School of Computer Science, University College Dublin, Ireland \\
$^2$ Department of Computer Science, Birkbeck, University of London, UK \\
\texttt{gechuan.zhang@ucdconnect.ie} \\
\texttt{p.nulty@bbk.ac.uk} \\
\texttt{david.lillis@ucd.ie}
}

\begin{document}
\maketitle
\begin{abstract}
The contextual word embedding model, BERT, has proved its ability on downstream tasks with limited quantities of annotated data. BERT and its variants help to reduce the burden of complex annotation work in many interdisciplinary research areas, for example, legal argument mining in digital humanities. Argument mining aims to develop text analysis tools that can automatically retrieve arguments and identify relationships between argumentation clauses. Since argumentation is one of the key aspects of case law, argument mining tools for legal texts are applicable to both academic and non-academic legal research. Domain-specific BERT variants (pre-trained with corpora from a particular background) have also achieved strong performance in many tasks. To our knowledge, previous machine learning studies of argument mining on judicial case law still heavily rely on statistical models. In this paper, we provide a broad study of both classic and contextual embedding models and their performance on practical case law from the European Court of Human Rights (ECHR). During our study, we also explore a number of neural networks when being combined with different embeddings. Our experiments provide a comprehensive overview of a variety of approaches to the legal argument mining task. We conclude that domain pre-trained transformer models have great potential in this area, although traditional embeddings can also achieve strong performance when combined with additional neural network layers.
\end{abstract}

\section{Introduction}
\label{sec:introduction}
Interdisciplinary research like digital humanities has been one of the most important trends in Natural Language Processing (NLP) research, where machine-based methods and tools are developed to automatically analyse texts from traditional humanities. Compared to general text analysis, in digital humanities many advanced approaches such as neural networks are still under-explored due to a lack of annotated datasets, which is a requirement of many supervised machine learning approaches. The complexity and labour cost required to produce new corpora constitute a major barrier in many areas of digital humanities research~\cite{Zhang2021}. Our specific research focus of legal argumentation mining is no exception in this regard.

The argument, a series of statements intended to determine the degree of truth for another statement, is one of the most important language structures used in the law. The ultimate goal of argument mining is to automatically identify arguments as well as their reasoning relations from texts~\citep{mochales2011argumentation}. For legal argument mining, creating corpora that can be used to develop automated systems requires case law to be annotated in a way that identifies the argumentation components and the relationship between them. This implies that the annotators should have sufficient related knowledge, and so the annotation process usually requires expert legal professionals such as lawyers or law school students. Although this legal Artificial Intelligence (AI) field has attracted much attention, there has still been a lack of development and successful deployment of applications, in part due to the dearth of comprehensive corpora for research.

Meanwhile, large pre-trained transformer models like BERT~\citep{devlin2019bert} have achieved outstanding performance on downstream tasks, even where only a limited amount of annotations are available. This has inspired a series of research studies on legal AI~\citep{chalkidis2019neural, reimers2019classification}. The pre-training plus fine-tuning strategy has improved both efficiency and performance when applying BERT. In particular, the model is first trained on a large dataset, then fine-tuned on the downstream tasks. The initial BERT model was trained on general corpus, which has inspired researchers to pre-train the model with different domain-specific corpora, for example legal texts~\citep{chalkidis2020legal}. Recent studies of BERT-base transformers pre-trained with legal corpora have displayed better performance on several legal AI tasks~\citep{xu2021accounting, silveira2021topic}. To gain insights into the improvements acheived by BERT-based models on legal argument mining, in this paper we compare the performance from two groups of embedding models: four BERT-based transformers pre-trained with legal texts, and two non-BERT embedding models. We also explore the enhancement of classic NLP neural networks on argument mining tasks.

Section~\ref{sec:related work} discusses the general background of argument mining and the original BERT model as well as introducing the domain pre-trained BERT variants and neural networks used in our experiments. Section~\ref{sec:dataset} provides details of the European Court of Human Rights (ECHR) case law corpus for argument mining. Section~\ref{sec:experiments} contains our experiment design and model implementation for argument extraction and relation prediction. We analyse the results in Section~\ref{sec:results} and conclude our work in Section~\ref{sec:conclusion}, along with a discussion of potential future work.

\section{Related Work} \label{sec:related work}
\subsection{Argument Mining} \label{sec:argument_mining_related}
As mentioned in Section~\ref{sec:introduction}, argument mining aims to automatically retrieve arguments and their related information from human language texts. Its interdisciplinary background makes argument mining a high-level research question in NLP, which is usually formalised as having two stages: \emph{argument extraction} and \emph{relation prediction}~\citep{cabrio2018five}. The first stage, argument extraction, aims to shrink the scope of argumentative texts (texts containing argument information) by filtering out unrelated parts and therefore focusing only on those sections of text that are argumentative. During the second stage of relation prediction, the relations between identified argumentative texts are predicted. Predicting the inner relations between argument components, or the outer relations between individual arguments, or both, depends on the practical requirements and the annotation scheme. Here, we focus on the inner structure of arguments, which includes identifying argument components and the relations between them.

To facilitate reasoning about arguments, a computational model of argument is needed. These are generally divided into two major categories: structural argumentation models and abstract argumentation models. The abstract argumentation model, also known as argumentation frameworks in~\cite{dung1995acceptability}, regards arguments themselves to be the elementary units, without additional internal structures. Nevertheless, the complexity of legal texts requires such internal argument structures, which leads to structural argumentation models becoming the main approaches when annotating legal corpora. A structural argumentation model usually consists of components and relations. Argument components are the elementary units, which are usually defined and annotated with their semantic role (e.g., premise, conclusion). Argument relations are the reasoning connections (e.g., support, defeat) between argument components (internal) and between individual arguments (external). Our experiment corpus currently focuses on internal argument relations, in particular the support relation between premise and conclusion.

One of the representative annotation standards for legal texts is the~\citet{walton2009argumentation} model. This is a tripartite structural argumentation model: a set of \emph{premises} that contain the evidence or reasons for supporting an argument, a \emph{conclusion} which is the stance and the central component of an argument, and the \emph{inference} from the set of premises to the conclusion. The high generalisability of Walton model makes it suitable for various contexts, for example, ECHR case law.

\subsection{BERT-based Models} \label{sec:bert transformer}

In order to develop tools for text analysis, models like word embeddings are applied to express human language features as computational vectors. As an advanced word embedding model, BERT (Bidirectional Encoder Representations from Transformers)~\citep{devlin2019bert} has achieved leading performance on several NLP areas, including legal text processing~\citep{chalkidis2019neural, reimers2019classification, poudyal2020echr}. BERT is a contextual word embedding model extracting text features with a deep transformer architecture~\citep{vaswani2017attention}. The complete training procedure of BERT is a two-step process. First, the model is trained on a large roughly labelled corpus using self-supervised learning methods. Next, when being adapted to downstream tasks, the model is further trained to fine-tune its weight-matrix with a well-annotated corpus that is usually much smaller. 

\subsubsection{BERT Pre-train Strategy}
A 16GB English corpus collected from online books \citep{zhu2015aligning} and Wikipedia were used for the pre-training of the original BERT model, BERT$_{base}$. The self-supervised learning process during the pre-training procedure has two objectives, masked language modelling (MLM) and next sentence prediction (NSP). Together, the pre-train process aims to enhance the model for deep bidirectional representations as well as sentence relationship understanding~\citep{devlin2019bert}.

\subsubsection{Legal Domain Pre-trained Models}
\label{sec:legal berts}
Legal language is considered to be a unique writing system which differs from generic text materials. Several researchers have explored whether using domain-specific pre-training corpora can enhance the performance of the BERT-base transformer for downstream tasks from the same domain~\citep{alsentzer2019publicly, beltagy2019scibert, lee2020biobert}. In our case, we focus on the BERT-base models pre-trained with legal text materials. To simplify the description, we use \emph{legal BERT models} as the BERT$_{base}$ variants which are pre-trained on text materials from the legal field. Previous legal text processing studies \citep{elwany2019bert, chalkidis2020legal, zhong2020does, zhong2020jec, zheng2021does} have demonstrated the improvements given by legal BERT models. In our experiment, we selected two groups of legal BERT models, the \emph{LEGAL-BERT Family} and the \emph{Harvard Legal-BERT Variants}.

\textbf{LEGAL-BERT Family} includes a series of BERT-based models using English legal texts for pre-training. Among them, we select two models: Legal-BERT$_{base}$ and Legal-BERT$_{echr}$. The total amount of pre-training text data collected for the LEGAL-BERT Family is 11.5GB, which covers several groups of legal documents, including EU and UK legislation, US court cases and contracts. \citet{chalkidis2020legal} explored two different domain-adaptation methods: 1) pre-training from scratch, 2) further pre-training. Legal-BERT$_{base}$ is pre-trained from scratch on the 11.5GB legal text collection, while Legal-BERT$_{echr}$ is further pre-trained with 0.5GB of ECHR case documents collected from the HUDOC\footnote{\url{https://hudoc.echr.coe.int/eng}} dataset.

\textbf{Harvard Legal-BERT Variants} are two legal BERT models pre-trained with judicial texts from the US court. In particular, \citet{zheng2021does} collected 37GB of text from the Harvard Law case corpus\footnote{\url{https://case.law/}}.
We name the two variants as Legal-BERT$_{harv}$ and Custom Legal-BERT$_{harv}$. Similar to~\citep{chalkidis2020legal}, \citet{zheng2021does} also trained their legal BERT models with different adaptation methods. Custom Legal-BERT$_{harv}$ is pre-trained from scratch using a custom vocabulary. Legal-BERT$_{harv}$ is further pre-trained with the same Harvard Law case corpus. Based on the corpus characteristics, they made adjustments to their pre-training objectives by using whole-word MLM and added regular expressions to ensure complete legal citations during the NSP.

\subsection{Other Approaches} \label{sec:other_approaches}

Apart from the set of embedding models pre-trained from BERT$_{base}$, another typical contextual word embedding model used in legal text processing is ELMo. In addition to the contextual word embedding models, GloVe, the traditional word embedding model, is still considered as a powerful feature extractor on legal texts.

\textbf{ELMo.} Unlike BERT with its deep transformer architecture, ELMo (Embedding from Language Models) \citep{peters2018deep} gains contextual word representations from a bidirectional structure. It consists of a character-based convolutional neural network (CNN) and two bidirectional long short-term memory (BiLSTM) layers.

\textbf{GloVe.} \citet{pennington2014glove} developed this unsupervised learning algorithm for obtaining word vector representations, pre-trained and stored as a dictionary-type matrix. The GloVe word vector representations exhibit linear substructures of the word vector space.

When designing legal text analysis models, after word embeddings, extra neural networks are typically applied to further extract the feature representation from embeddings to higher-level encoded vectors. Several legal text processing studies have combined word embedding models with layers such as classic BiLSTM, CNN, and ResNet.

\textbf{BiLSTM.} The long short-term memory (LSTM) is a recurrent neural network (RNN) using memory gates and hidden states to extract and calculate the text features sequentially. BiLSTM (bidirectional LSTM) is a variant LSTM network which summarises information from both directions. BiLSTM also works as a significant module in ELMo embeddings for generating contextual vector representations from input texts. The legal text processing experiment presented in in \citep{zheng2021does} used BiLSTM network as the baseline model.

\textbf{CNN.}
The convolutional neural network (CNN) is one of the basic machine learning models, which utilises convolutional filters to extract the text features. This is a basic model in legal text processing. \citep{zhong2019automatic} developed a pipeline for legal decision summarisation with a CNN classifier. Similarly, \citep{xu2021toward} applied a CNN model for legal text classification,and also achieved good results through connecting CNN and BERT embeddings.

\textbf{ResNet.}
The residual network \citep{he2016deep} is a classic architecture with special shortcuts that connect neurons belonging to distant layers, which is different from the traditional feed-forward networks. The shortcut design provides ResNet with high efficiency when calculating deep networks with multiple layers. The study of argumentative link prediction in \citep{galassi2018argumentative} presented a combination model of ResNet and GloVe embeddings, which performed well.

\section{ECHR Dataset}
\label{sec:dataset}
For our experiments, we choose the argument mining corpus annotated on ECHR case-laws from the HUDOC database: an open-source database that has been commonly used for legal AI studies such as court decision event extraction~\citep{filtz2020events}, judicial decision prediction~\citep{chalkidis2019neural, medvedeva2020using}, and legal argument mining~\citep{mochales2011argumentation, teruel2018increasing, poudyal2020echr}. Since the early stage of argument mining research~\citep{mochales2011argumentation}, ECHR case law has been used as a practical application scenario. Detailed information of the argumentation structure in ECHR case law has been provided in \citep{mochales2008study}. In our experiment, we used the ECHR argument mining corpus (ECHR-AM), annotated and open-sourced in \citep{poudyal2020echr}\footnote{\url{http://www.di.uevora.pt/~pq/echr/}}.

\begin{figure}[htbp]
\centering
\includegraphics[width=0.8\linewidth]{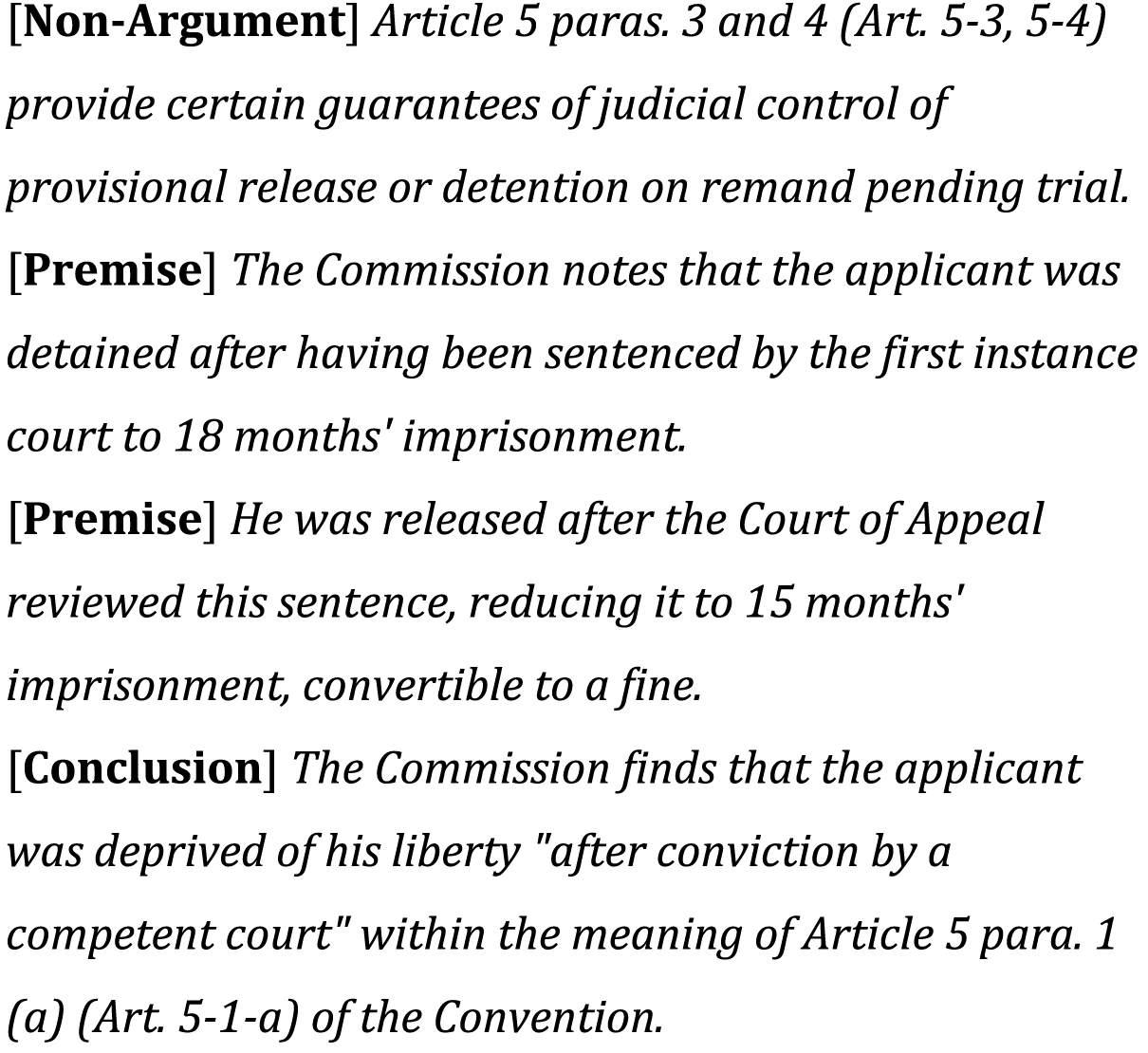}
\caption{Annotation Example of the ECHR Argument Mining Corpus}
\label{fig:pipeline}
\end{figure}

The ECHR-AM corpus contains text from 42 cases (20 decisions and 22 judgements) including approximately 290,000 words. The ECHR-AM annotation scheme is designed based on Walton's premise/conclusion model. The legal text is first segmented into sentence-level clauses then further annotated into three groups: premises, conclusions, and non-argument clauses. The premises and conclusions are argument components defined by the argumentation model. The amount of arguments is quite unbalanced between documents, ranging from a minimum of 4 arguments (8 premises and 4 conclusions) to 50 arguments (147 premises and 50 conclusions). As a result, we use document-level train-test splitting during our experiment, to verify the model’s performance in practical situations.

\section{Experiments} \label{sec:experiments}
The aim of our experiments is to explore which language models, word embeddings and machine learning techniques are most suited to the task of legal argument mining. The following sections describe how the legal argument mining task is structured, the system architecture used for the experiments, and other specific details of the experiment setup.

\subsection{Legal Argument Mining Tasks} \label{sec:Legal Argument Mining Tasks}
Our implementation of the legal argument mining task primarily follows the previous experiments in~\citep{poudyal2020echr}, where the entire argument mining process is separated into three distinct tasks. To align with this approach, we have structured our work as a pipeline that consists of: argument clause recognition, argument relation mining, and argument component classification (including the classification of both premises and conclusions).

\subsubsection{Argument Clause Recognition}
In the first task of our argument mining pipeline, we aim to shrink the range of argument information in case documents by filtering out non-argument clauses. We formalise this task into a binary classification problem, and train the model to detect argument clauses, which are clauses containing argumentation information (either a  premise or a conclusion).

\subsubsection{Argument Relation Mining} \label{sec:Argument Relation Mining}

Since the annotation scheme of the ECHR-AM corpus only includes the inner relations within arguments, we focus on identifying the support relations from the set of premises to the conclusion, which have been defined as the inference in Walton model (see Section~\ref{sec:argument_mining_related}). This task of relation mining is considered as the bottleneck not only with legal texts, but in the whole research field of argument mining~\citep{mochales2011argumentation, poudyal2019using, poudyal2020echr}. The goal of this task is to group clauses into different arguments before further identifying their labels or functions as specific argument components.

Due to the fact that an argument clause may belong to multiple arguments, we manage this task with the same implementation method as in~\citep{poudyal2020echr}, where they modelled this as a as clause-pair relation prediction task. A classifier model is used to predict whether a pair of input clauses (already identified as argumentative from the previous step) are from the same argument or not. In particular, we assume all the argument clauses have been successfully detected from previous task. We first order the argument clauses within the same document into a sequence, then use a fixed-size sliding window (size = 5) to generate the clause-pair inputs, which are further classified into related or non-related groups.

\subsubsection{Argument Component Classification} \label{sec:Argument Component Classification}
Following the previous design of argument component classification~\citep{mochales2011argumentation, poudyal2020echr}, we treat this task as two separate text classification problems. Since the practical situation that an argument clause may act as premise in one argument and also be the conclusion in another, we apply two individual classifiers for premise recognition and conclusion recognition respectively.

\subsection{Method Architecture}
The general classifier in our experiment contains two parts: an embedding module for handling the text input, and an encoder module for compressing the feature vectors into categories output. By adapting different models into our classifier structure, this work employs a series of combinations which differ as to the way in which the text is embedded (BERT, ELMo, GloVe), and how the embedding vectors are encoded as output (BiLSTM, CNN, ResNet). Since our experiment design initially compares the performance of various embedding models, we choose the classification results from embeddings as the baselines in our research.

We include four domain-specific BERT models selected from two groups (see Section~\ref{sec:legal berts}). Besides the transformers, we use ELMo as the another contextual embedding model. The final ELMo embedding is obtained by the three mixed representation layers from the pre-trained 5.5B ELMo model. To compare with contextual embedding models, we refer to the previous study~\citep{galassi2021multi} of using GloVe word embeddings. We use the GloVe pre-trained vocabulary (840B) and turn input clauses into 300-dimensional embeddings. For each baseline model, we generate embeddings from the input segmented clauses, and add a simple classifier head containing a dropout layer (dropout rate = 0.1) and a liner layer (for the final output).

Apart from the baselines provided by both contextual and traditional word embeddings, we study the models' classification performance with extra encoder modules. First, we apply a BiLSTM (100-dimensional) layer to test the enhancement of a recurrent neural network. Second, we implement a four-layer CNN model with 100 convolution filters in each layer and 4 kernel sizes (3, 4, 5, 6). Each convolution layer in the model is activated by a ReLU function and stacked with one max pooling layer. Finally, by implementing the residual bottleneck block, we explore the use of ResNet in our experiment as well. We use a three-layer ResNet, composed by two layers with 64 filters and one layer with 256 filters.

\subsection{Experimental Setup}
Following the previous setup given by~\citet{poudyal2020echr}, we use 80\% of the legal documents (34 documents) for training and the remaining 20\% (8 documents) for testing. We use k-fold (k=5) cross validation for fine-tuning and selecting the model for testing, where each fold uses 20\% of the training set for validation purposes. Before input into the embedding model, we pad the pre-processed token sequences with the same length (250 for single clause inputs, 500 for clause pair inputs, heuristically). Cross-entropy loss function and Adam optimiser (initial learning rate = 2e-5) are used for optimising the model. An early-stopping strategy (patience = 5) is used to stop the training procedure when the validation loss has not been decreased. For pre-trained embedding models with deep network layers, the tuning processes are shorter than randomly initialised networks~\citep{devlin2019bert, liu2019roberta}. Considering the experimental setup in previous studies~\citep{chalkidis2020legal, bonifacio2020study, Wang2020a}, we train the BERT-base models with maximum 10 epochs, ELMo-base models with maximum 20 epochs, and GloVe-base models with maximum 150 epochs. In our experiment, we find most models stop early before reaching the maximum number of epochs. Due to the unbalanced distribution of the ECHR-AM dataset, we select the weighted scikit-learn evaluation metrics of precision, recall and F1 measure in our experiments, which account for label imbalance during calculation \citep{scikit-learn}. We perform five runs for each model and report the mean evaluation scores.

\section{Results and Discussion} \label{sec:results}

This section presents the results\footnote{\url{https://github.com/LegalAM/JDMDH-2022-ECHR}} of our experiments for each of the three phases of the argument mining pipeline.

\subsection{Argument Clause Recognition Results}
Previous studies of the argumentation structures in ECHR case-law suggest that the majority of the argument information is concentrated within specific sections (e.g., ``AS TO THE LAW/THE LAW'' sections)~\citep{mochales2008study, poudyal2020echr}. Besides, to align with previous experiments in~\citep{poudyal2020echr}, we employ two search scopes for argument clauses: a small one without contents before the ``AS TO THE LAW/THE LAW'' section; and the complete one that uses the entire document contents. In Table~\ref{tab:result1}, the first three columns indicate how well the model performs on the specific-section search scope, and the rest are results of argument clause recognition on full-text ECHR cases.

\begin{table*}[htbp]
\small
\centering
\caption{Weighted precision (P), recall (R), F1 measurement for the argument clause recognition task on the specific section scope and the whole document scope (sec = specific section, full = full text, C = Custom, underlines for baseline model scores, bold texts for best scores in each sub-task experiment, stars for best scores in model combination groups).}
\label{tab:result1}
    \begin{tabular}{l|ccc|ccc}
    \hline
    \textbf{Model Combination} & \textbf{P-sec} & \textbf{R-sec} & \textbf{F1-sec} & \textbf{P-full} & \textbf{R-full} & \textbf{F1-full}\\
    \hline
    GloVe & .519 & .614 & \underline{.488} & .777 & .789 & \underline{.747}\\
    GloVe+bilstm & .423 & .556 & .424 & .752 & .770 & .740\\
    GloVe+cnn & .803* & .787* & .778* & \textbf{.910}* & \textbf{.911}* & \textbf{.908}*\\
    GloVe+resnet & .747 & .738 & .723 & .831 & .839 & .826\\
    \hline
    ELMo & .748 & .727 & \underline{.698} & .793 & .802 & \underline{.787}\\
    ELMo+bilstm & .710 & .659 & .604 & .773 & .784 & .767\\
    ELMo+cnn & .784* & .764* & .750 & .855 & .856 & .849\\
    ELMo+resnet & .772 & .762 & .752* & .860* & .864* & .856*\\
    \hline\hline
    Legal-BERT$_{base}$ & .788 & .779 & \underline{.771} & .876 & .885 & \underline{.877}\\
    Legal-BERT$_{base}$+bilstm & .801* & .800* & .794* & .874 & .871 & .871\\
    Legal-BERT$_{base}$+cnn & .796 & .785 & .776 & .893* & .891* & .891*\\
    Legal-BERT$_{base}$+resnet & .730 & .736 & .723 & .868 & .875 & .865\\
    \hline
    Legal-BERT$_{echr}$ & .814* & .806* & \underline{.800} & .905* & .902* & \underline{.902}*\\
    Legal-BERT$_{echr}$+bilstm & .803 & .793 & .788 & .895 & .891 & .891\\
    Legal-BERT$_{echr}$+cnn & .776 & .761 & .750 & .899 & .899 & .898\\
    Legal-BERT$_{echr}$+resnet & .807 & .804 & .803* & .877 & .875 & .874\\
    \hline
    C-Legal-BERT$_{harv}$ & .795 & .792 & \underline{.787} & .860 & .861 & \underline{.858}\\
    C-Legal-BERT$_{harv}$+bilstm & .801 & .798 & .791 & .889* & .886* & .887*\\
    C-Legal-BERT$_{harv}$+cnn & \textbf{.825}* & \textbf{.819}* & \textbf{.817}* & .873 & .873 & .872\\
    C-Legal-BERT$_{harv}$+resnet & .781 & .776 & .774 & .866 & .865 & .864\\
    \hline
    Legal-BERT$_{harv}$ & .780 & .778 & \underline{.769} & .862 & .864 & \underline{.860}\\
    Legal-BERT$_{harv}$+bilstm & .793 & .788 & .780 & .876 & .876 & .875*\\
    Legal-BERT$_{harv}$+cnn & .814* & .800* & .792* & .878* & .876* & .875*\\
    Legal-BERT$_{harv}$+resnet & .770 & .769 & .763 & .868 & .870 & .866\\
    \hline
    \end{tabular}
\end{table*}

Here we start by presenting and analysing the results of argument clause recognition in specific sections. For the non-BERT embedding baselines given by GloVe and ELMo, the weighted F1 measure from ELMo is much higher (0.698~vs.~0.488). The additional CNN encoder, on the other hand, substantially improved the GloVe-based model’s performance. When applied together, GloVe+cnn reached outstanding evaluation scores (precision = 0.803, recall = 0.787, and F1 = 0.778), which are the highest among all the GloVe-based models. CNN also enhanced the performance of ELMo embeddings, by increasing the F1 score (0.698~vs.~0.750). The ELMo+cnn combination also obtained the ELMo group-wise greatest precision (0.784) and recall (0.764). Meanwhile, the ResNet encoder greatly increased both non-BERT models’ performance of identifying argumentative information, where GloVe+resnet reached better evaluation scores than its baseline (precision: 0.747~vs.~0.519, recall: 0.738~vs.~0.614, and F1: 0.723~vs.~0.488), and ELMo+resnet gained its group-wise maximum F1 value (0.752). It is obvious that multi-layered CNN and ResNet have the good ability to compress text feature vectors from embeddings. Besides, connecting GloVe embeddings with deeper networks like CNN can substantially improve the classifier's ability.

The results of argument clause recognition generally aligned to previous research of applying domain-specific pre-trained BERT models on the legal argument mining downstream task~\citep{xu2021accounting}. Among all four BERT embedding models, Legal-BERT$_{echr}$ reached the leading F1 score (0.800), compared to other baseline F1 results (Legal-BERT$_{base}$ F1 = 0.771, Legal-BERT$_{harv}$ F1 = 0.769, and Custom Legal-BERT$_{harv}$ F1 = 0.787), along with remarkable precision (0.814) and recall (0.806). When applied with the ResNet encoder, the performance of Legal-BERT$_{echr}$+resnet improved only very slightly when compared to its baseline F1 measure (0.803~vs.~0.800). Legal-BERT$_{base}$ was strengthened by adding extra BiLSTM layer as its encoder. All three evaluation scores of Legal-BERT$_{base}$+bilstm improved about 2\%. For the two Harvard Legal-BERT variants, when using CNN as the extra encoder module, both models reached their group-wise best performance. All three scores approximately increased 3\% in both evaluation tests of Custom Legal-BERT$_{harv}$ and Legal-BERT$_{harv}$ with the CNN encoder. The evaluation scores obtained by Custom Legal-BERT$_{harv}$+cnn are also the highest among all legal BERT models (precision = 0.825, recall = 0.819, and F1 = 0.817). Besides, like Legal-BERT$_{base}$, the BiLSTM layer also made both Harvard Legal-BERT models slightly better.

When enlarge the search scope to entire ECHR documents, the gap between the GloVe and ELMo baselines of weighted F1 shrank (0.747~vs.~0.787), where ELMo yet maintained its higher score. Similar to the previous argument clause recognition task within sections, both GloVe+cnn and ELMo+resnet reached their own group-wise best performance, among which the evaluation results of GloVe+cnn (precision = 0.910, recall = 0.911, and F1 = 0.908) were even slightly better than those of the legal BERT models. Apart from that, ResNet encoders reinforced the performance of both GloVe and ELMo (around 6\%). Among all BERT pre-train models, Legal-BERT$_{echr}$ kept the best evaluation results (precision = 0.905, recall = 0.902, and F1 = 0.902). Apart from GloVe and ELMo, CNN remained useful for pre-trained transformers, the weighted F1 scores of Legal-BERT$_{base}$, Legal-BERT$_{harv}$, and Custom Legal-BERT$_{harv}$ all increased. Likewise, adding the extra BiLSTM also improved both Harvard Legal-BERT variants. In general, BERT variants outperform both GloVe and Elmo as embedding models, which shows the strong ability of pre-trained transformers to distinguish argument information form general text in legal documents. Adding extra convolutional and residual networks sightly improved the performance of some legal BERT models, but magnified the performance of GloVe embedding greatly. We suggest that the transformer model (i.e., BERT) already has complex network layers which is why adding the extra encoder layer has less improvement compared to simple embedding layers like GloVe.

\subsection{Argument Relation Mining Results}
Table~\ref{tab:result2} shows the results of the argument relation mining subtask. After adapting with deeper networks, the GloVe embedding model has better performance when processing paired long text and predicting the relations of clause pairs. Both GloVe+bilstm and GloVe+resnet reached higher F1 scores compared to their baseline (0.568~vs.~0.506, and 0.574~vs.~0.506). CNN has the greatest influence among all neural network encoders, which increased all three evaluations from the GloVe baseline remarkably (precision: 0.732~vs.~0.562, recall: 0.729~vs.~0.631, and F1: 0.703~vs.~0.506). When combined with additional BiLSTM and ResNet encoders, ELMo’s performance was undermined, the F1 score decreased. Yet, ELMo+cnn continued its improved performance and reached the group-wise best F1 score (0.722). On the other hand, when applying different BERT models on the argument relation mining, Legal-BERT$_{echr}$ maintained its good ability on sentence-pair classification and obtained the best evaluation baseline among all pre-trained transformers (precision = 0.775, recall = 0.772, and F1 = 0.765). Besides, the BiLSTM layer slightly improved the whole LEGAL-BERT family, which increased the F1 score of Legal-BERT$_{base}$ and Legal-BERT$_{echr}$ (0.727~vs.~0.757, 0.765~vs.~0.771). Compared to GloVe and ELMo, the legal BERT models had conducted great results when mining argument relations even without an extra encoder.

\begin{table*}[htbp]
\small
\centering
\caption{Weighted precision (P), recall (R), F1 measurement for the argument relation mining task (C = Custom, underlines for baseline model scores, bold texts for best scores in this task experiment, stars for best scores in model combination groups).}
\label{tab:result2}
    \begin{tabular}{lccc}
    \hline
    \textbf{Model Combination} & \textbf{P} & \textbf{R} & \textbf{F1}\\
    \hline
    GloVe & .562 & .631 & \underline{.506}\\
    GloVe+bilstm & .590 & .644 & .568\\
    GloVe+cnn & .732* & .729* & .703*\\
    GloVe+resnet & .639 & .632 & .574\\
    \hline
    ELMo & .709 & .713 & \underline{.683}\\
    ELMo+bilstm & .673 & .658 & .610\\
    ELMo+cnn & .757* & .740* & .722*\\
    ELMo+resnet & .711 & .707 & .680\\
    \hline\hline
    Legal-BERT$_{base}$ & .744 & .738 & \underline{.727}\\
    Legal-BERT$_{base}$+bilstm & .768* & .770* & .757*\\
    Legal-BERT$_{base}$+cnn & .750 & .750 & .742\\
    Legal-BERT$_{base}$+resnet & .737 & .743 & .730\\
    \hline
    Legal-BERT$_{echr}$ & .775 & .772 & \underline{.765}\\
    Legal-BERT$_{echr}$+bilstm & \textbf{.779}* & \textbf{.776}* & \textbf{.771}*\\
    Legal-BERT$_{echr}$+cnn & .769 & .766 & .761\\
    Legal-BERT$_{echr}$+resnet & .758 & .757 & .750\\
    \hline
    C-Legal-BERT$_{harv}$ & .734 & .740 & \underline{.728}\\
    C-Legal-BERT$_{harv}$+bilstm & .731 & .730 & .720\\
    C-Legal-BERT$_{harv}$+cnn & .764* & .768* & .757*\\
    C-Legal-BERT$_{harv}$+resnet & .724 & .738 & .723\\
    \hline
    Legal-BERT$_{harv}$ & .762* & .762* & \underline{.756}*\\
    Legal-BERT$_{harv}$+bilstm & .735 & .746 & .732\\
    Legal-BERT$_{harv}$+cnn & .740 & .741 & .735\\
    Legal-BERT$_{harv}$+resnet & .754 & .755 & .746\\
    \hline
    \end{tabular}
\end{table*}

\subsection{Argument Component Classification Results}
As mentioned in Section~\ref{sec:Legal Argument Mining Tasks}, the practical argumentation in legal texts include overlaps between individual arguments where a clause can be a premise in one argument and the conclusion in another. Since each argument clause may require multiple component labels, like~\citep{mochales2011argumentation, poudyal2020echr}, we maintain the separation between the two individual sub-tasks of classifying premises/non-premises and conclusions/non-conclusions. The results of the two argument component classification sub-tasks are recorded in Table~\ref{tab:result3}. The first three columns display the weighted precision, recall, F1 measure for the classification of conclusions; the next three columns display the same evaluation scores for the classification of premises; the last column of the table is the average F1 score which stands for the average of class-wise (premise/conclusion) F1 scores.

\begin{table*}[htbp]
\small
\centering
\caption{Weighted precision (P), recall (R), F1 measurement for the argument component (premise/conclusion) classification task (con = conclusion, pre = premise, C = Custom, underlines for baseline model scores, bold texts for best scores in each sub-task experiment, stars for best scores in model combination groups).}
\label{tab:result3}
    \begin{tabular}{l|ccc|ccc|c}
    \hline
    \textbf{Model Combination} & \textbf{P-con} & \textbf{R-con} & \textbf{F1-con} & \textbf{P-pre} & \textbf{R-pre} & \textbf{F1-pre} & \textbf{avg-F1}\\
    \hline
    GloVe & .514 & .717 & \underline{.598} & .518 & .720 & \underline{.603} & \underline{.601}\\
    GloVe+bilstm & .527 & .726 & .610 & .511 & .714 & .595 & .603\\
    GloVe+cnn & .817* & .814* & .790* & .824* & .820* & .800* & .795*\\
    GloVe+resnet & .774 & .775 & .727 & .748 & .757 & .702 & .714\\
    \hline
    ELMo & .770 & .774 & \underline{.738} & .782 & .788 & \underline{.763} & \underline{.750}\\
    ELMo+bilstm & .749 & .766 & .716 & .761 & .768 & .709 & .713\\
    ELMo+cnn & .799* & .806 & .791* & .833* & .832* & .818* & .805*\\
    ELMo+resnet & .799* & .807* & .788 & .808 & .814 & .797 & .792\\
    \hline\hline
    Legal-BERT$_{base}$ & .843 & .843 & \underline{.837} & .833 & .838* & \underline{.829} & \underline{.833}\\
    Legal-BERT$_{base}$+bilstm & .845 & .847 & .844* & .834* & .836 & .832* & .838*\\
    Legal-BERT$_{base}$+cnn & \textbf{.846}* & \textbf{.849}* & .842 & .831 & .833 & .825 & .834\\
    Legal-BERT$_{base}$+resnet & .826 & .830 & .817 & .827 & .834 & .821 & .819\\
    \hline
    Legal-BERT$_{echr}$ & .840 & .841 & \underline{.837} & .828 & .831 & \underline{.825} & \underline{.831}\\
    Legal-BERT$_{echr}$+bilstm & .842* & .843* & .840* & \textbf{.860}* & \textbf{.862}* & \textbf{.859}* & \textbf{.850}*\\
    Legal-BERT$_{echr}$+cnn & .834 & .833 & .829 & .836 & .838 & .835 & .832\\
    Legal-BERT$_{echr}$+resnet & .832 & .836 & .829 & .831 & .834 & .828 & .828\\
    \hline
    C-Legal-BERT$_{harv}$ & .838* & .840* & \underline{.836}* & .834 & .837 & \underline{.834} & \underline{.835}*\\
    C-Legal-BERT$_{harv}$+bilstm & .826 & .829 & .822 & .839* & .843* & .838* & .830\\
    C-Legal-BERT$_{harv}$+cnn & .833 & .837 & .831 & .830 & .832 & .822 & .826\\
    C-Legal-BERT$_{harv}$+resnet & .833 & .836 & .832 & .823 & .828 & .820 & .826\\
    \hline
    Legal-BERT$_{harv}$ & .845* & \textbf{.849}* & \underline{\textbf{.845}}* & .848 & .850 & \underline{.847} & \underline{.846}*\\
    Legal-BERT$_{harv}$+bilstm & .830 & .833 & .824 & .835 & .838 & .833 & .829\\
    Legal-BERT$_{harv}$+cnn & .829 & .834 & .827 & .852* & .855* & .849* & .838\\
    Legal-BERT$_{harv}$+resnet & .830 & .835 & .824 & .849 & .852 & .847 & .835\\
    \hline
    \end{tabular}
\end{table*}

When identifying conclusions, a similar pattern to the previous tasks emerges in terms of the performance of the two non-BERT embedding baselines. Again, the weighted F1 score of ELMo greatly outperforms the F1 score of GloVe (0.738~vs.~0.598). By adding the extra neural networks, the performance of the GloVe-base model was greatly improved, and the gap between these two models’ evaluation scores was reduced. For the weighted F1 score, GloVe+cnn and ELMo+cnn were almost equal (0.790~vs.~0.791), each of which is also the best F1 in its group. The combination of GloVe+cnn also gained precision (0.817) and recall (0.814) scores which respectively exceed the best ELMo-based scores (from ELMo+resnet, precision = 0.799, recall = 0.807). When using BERT-base transformers, the general classification results of conclusion are better than GloVe and ELMo embeddings, as expected. Legal-BERT$_{harv}$ outperforms all the other pre-trained BERT variants with the top F1 measure (0.845). It is interesting that adding extra layers did not improve the ability of the Harvard Legal BERT models to identify conclusions. We suggest this is likely caused by the limited amount of positive data in the test set. On the other hand, BiLSTM improved the classification results of conclusions for both pre-trained models from the Legal-BERT family. Both Legal-BERT$_{base}$+bilstm and Legal-BERT$_{echr}$+bilstm obtained their group-wise highest F1 scores (0.844 and 0.840), in which the F1 of Legal-BERT$_{base}$ is almost the same as the best (0.845 from Legal-BERT$_{harv}$) within the conclusion classification task.

The performance of both GloVe and ELMo embeddings on the identification of premises are aligned to their results on the classification of conclusions, but generally better. The weighted F1 score of GloVe increased (0.603 vs 0.598) when switching the classification target from conclusion to premise. Similarly, ELMo’s performance slightly increased when identifying premises (0.763 vs 0.738). CNN and ResNet maintained their ability to enhance the performance of both non-BERT embedding models. For the CNN encoder, GloVe+cnn upgraded the evaluation F1 score almost 20\% from the GloVe embedding baseline (0.800~vs.~0.603). Elmo+cnn also reached its group-wise best scores (precision = 0.833, recall = 0.832, and F1 = 0.818). For the ResNet encoder, the improvement for ELMo was not as significant as for GloVe, but still helped ELMo+resnet earned better results compared to the baseline (precision: 0.808~vs.~0.782, recall: 0.814~vs.~0.788, and F1: 0.797~vs.~0.763). The average performance of legal BERT models maintained higher scores than non-BERT embeddings. Legal-BERT$_{harv}$, which has the top F1 score (0.845) when identifying conclusions, reached the highest baseline F1 (0.847) again when detecting premises. By using the BiLSTM encoder, both models from the LEGAL-BERT family were augmented. The evaluation scores given by Legal-BERT$_{echr}$+bilstm are the highest in the premise classification task (precision = 0.860, recall = 0.862, and F1 = 0.859). Legal-BERT$_{base}$+bilstm also reached its group-wise best performance in both sub-tasks of argument component classification. It is noteworthy that the BiLSTM network raised the performance of Legal-BERT$_{echr}$ from the lowest BERT embedding baseline to the best classifier in the premise classification sub-task (0.825 vs 0.850). We use an extra evaluation score, which is the average value of both weighted F1 scores from the two sub-tasks (premise/conclusion) of argument component classification. When evaluating all the models using average F1, the CNN encoder again demonstrated its ability to enhance the performance of both GloVe and ELMo embeddings (0.795 vs 0.601, 0.805 vs 0.750). Although the Harvard BERT variants did not reach better performance with an additional encoder, the LEGAL-BERT family were enhanced with extra layer of BiLSTM, and Legal-BERT$_{echr}$+bilstm achieved the best average weighted F1 (0.850) among all transformer-based models.

\section{Conclusion and Future Work}
\label{sec:conclusion}

In this paper, we selected multiple legal BERT variants as well as other classic pre-trained embedding models, and provided a broad study of word embedding models' performance on legal argument mining. Our study currently focuses on practical case law from the European Court of Human Rights (ECHR). During our experiments on word embedding models, we also adapted a number of classic NLP neural networks. A comprehensive evaluation of these combinations in the context of argument mining had not been conducted to date, to our knowledge. Thus, our experiments  help to contribute to the set of baselines available to researchers going forward, as well as showing how these neural networks can enhance the performance of state-of-the-art transformer-base argument mining models.

Overall, legal BERT embeddings have better performance than ELMo and GloVe in most of the argument mining tasks. The strong performance of BERT models in the argument relation mining task suggests their great generalisability when handling long inputs of clause pairs. When applied with a BiLSTM network, the two models from the LEGAL-BERT family were enhanced for mining argument relations, which implies BiLSTM’s improvement when processing long sequential inputs. Domain pre-training also improves the text classification performance when BERT is applied on a small dataset with similar sentence contents (e.g., argument component classification tasks). The Legal-BERT$_{echr}$ model is pre-trained with a limited domain-specific corpus, while exhibiting outstanding performance in both the relation prediction and component classification tasks. It indicates that the characteristics of legal language are quite distinct from that of general English texts, and also leads us to a conclusion that domain-specific pre-training can work effectively on this type of interdisciplinary downstream tasks, with a special language context. It is logical to believe that this approach will be beneficial in other digital humanities applications besides the legal domain also. Besides, from our experiments, both convolutional-base networks (CNN and ResNet) have displayed their ability to enhance GloVe and ELMo embedding models. In particular, the combination of GloVe+cnn showcases its classification ability when detecting small groups of argument clauses from the full text of legal documents.

This work still follows the classic pipeline structure for argument mining system design, which inevitable includes the error propagation issue between its serial tasks. This brings difficulties for evaluation as well for its practical application outside the laboratory environment.. To solve this significant problem, several novel methods and implementation strategies (e.g., joint learning \citep{niculae2017argument}, multi-task learning strategy \citep{galassi2021multi}, and graph neural network \citep{ye2021end}) have made breakthroughs on other argument mining corpora. We argue that those techniques are potential solutions for the error propagation issue in legal argument mining. Moreover, adjusting input token length may improve the embedding models' performance when dealing with long input texts from legal documents~\citep{limsopatham2021effectively}. Adapting these methods is part of our future work for updating the current traditional mining process.

\bibliography{reference}
\bibliographystyle{acl_natbib}

\end{document}